\def\year{2019}\relax
\documentclass[letterpaper]{article} %DO NOT CHANGE THIS
\pdfoutput=1

\label{User-Define-Before-Begin}
\usepackage{authblk}
\label{User-Define-End}

\usepackage{aaai19}  %Required

\usepackage{times}  %Required
\usepackage{helvet}  %Required
\usepackage{courier}  %Required
\usepackage{url}  %Required
\usepackage{graphicx}  %Required
\label{User-Defined-Begin}

\newif\ifshowstruct
\showstructtrue
\showstructfalse % comment here and show the struct of paper

\ifshowstruct
	\newcommand{\tstructz}[1]{[#1] }
	\usepackage[UTF8]{ctex}%tmp
\else
	\newcommand{\tstructz}[1]{}
\fi

\usepackage{relsize}
\newcommand{\miss}[0]{ \mathsf o }
\newcommand{\W}[0]{ {\mathsf H} }
\newcommand{\ob}[0]{a}
\newcommand{\loaded}[1][ ]{available#1}
\newcommand{\etparam}[1][t]{,a_{#1};\zeta_{#1}}
\newcommand{\missparam}{\zeta} % !!!!!!!!!!!!!Remeber to check the exeperiment

\newif\ifshowtmp
\showtmptrue
\showtmpfalse % comment here and show the struct of paper

\ifshowtmp
	\newcommand{\ttt}[1]{\textcolor{red}{#1}}

\else
	\newcommand{\ttt}[1]{#1}

\fi

\newcommand{\ProposedMethod}[1][ ]{BI-PPO#1}
\newcommand{\fig}{Fig.~}
\newcommand{\new}[1]{{\color{black}{#1}}}
\usepackage{hyperref}
\hypersetup{
	colorlinks,
	linkcolor={blue},
	citecolor={blue},
	urlcolor={blue}
}
\usepackage{graphicx,subfigure}
\usepackage{caption}

\usepackage{multirow}
\usepackage{multicol}

\usepackage{array}
\newcolumntype{+}{>{\global\let\currentrowstyle\relax}}
\newcolumntype{Y}{>{\currentrowstyle}}
\newcommand{\rowstyle}[1]{\gdef\currentrowstyle{#1}%
	#1\ignorespaces
}
\usepackage{xcolor}
\usepackage{colortbl,booktabs}
\usepackage{color}

\usepackage{amsmath, amssymb, natbib, graphicx, url, mathtools }%algorithm2e
\usepackage{amsthm}
\theoremstyle{plain}

\usepackage{wrapfig}
\usepackage{algorithm,algorithmic}    
\usepackage{xr}

\usepackage{breqn}

\newcommand{\breakingcomma}{
  \begingroup\lccode`~=`,
  \lowercase{\endgroup\expandafter\def\expandafter~\expandafter{~\penalty0 }}
}
\usepackage{footnote}
\makesavenoteenv{tabular}
\makesavenoteenv{table}
\usepackage{stfloats}
\usepackage{pdfpages}
\label{User-Defined-End}

\begin{document}
\title{Robust Reinforcement Learning in POMDPs \\
with Incomplete and Noisy Observations}

\renewcommand*{\Authsep}{, }
\renewcommand*{\Authand}{, }
\renewcommand*{\Authands}{, }
\author[1]{Yuhui Wang}
\author[1]{Hao He}
\author[1*]{Xiaoyang Tan}
\affil[1]{Nanjing University of Aeronautics and Astronautics, China}
\affil[]{\textit {\{y.wang, hugo, x.tan\}@nuaa.edu.cn}}

\maketitle

\begin{abstract}
In real-world scenarios, the observation data for reinforcement learning with continuous control is commonly noisy and part of it may be dynamically missing over time, which violates the assumption of many current methods developed for this. We addressed the issue within the framework of partially observable Markov Decision Process (POMDP) using a model-based method, in which the transition model is estimated from the incomplete and noisy observations using a newly proposed surrogate loss function with local approximation, while the policy and value function is learned with the help of belief imputation. For the latter purpose, a generative model is constructed and is seamlessly incorporated into the belief updating procedure of POMDP, which enables robust execution even under a significant incompleteness and noise. The effectiveness of the proposed method is verified on a collection of benchmark tasks, showing that our approach outperforms several compared methods under various challenging scenarios.
\end{abstract}

\section{Introduction}\label{sec_intro}
	Significant progress has been made in reinforcement learning (RL) to solve a large number of tasks, such as Atari games \citep{mnih2015human}, board games \citep{silver2017mastering1,silver2017mastering} and robotic control \citep{Schulman2016HighDimensional}. On tasks like robotic control, the agent acquires full observation through sensors from the environment to make decisions. For example, a bipedal robot perceives data through sensors such as position sensor and velocity sensor. 
	However, in real-world scenarios, the agent could receive \emph{incomplete observation} data for some reasons, namely, part of the sensor data is missing. For example, any malfunction in the sensor, too much time of preprocessing, or intrinsically, the sensors' different sampling frequency from each other, could result in this issue \citep{jette1997reinforcement}.
	Furthermore, real-world applications usually involve \emph{noise} coming from sensors, non-deterministic actions and environment \citep{framling2004reinforcement}. 
%	For example, a quadruped robot could sustain noise from an imprecise sensor, an unstable actuator, and an uneven floor.
	Current RL systems are not robust against these incomplete and noisy data. For example, the Proximal Policy Optimization (PPO) algorithm, involving policy network and value network, which requires complete observation to output actions, and the learning process could be damaged by the noise.
	
%	\begin{wrapfigure}{R}{0.35\textwidth}
%	\begin{figure}
%		\centering
%%		\includegraphics[width=0.35\textwidth]{figs/fig_incomplete_observation.pdf}
%		\includegraphics[width=0.30\textwidth]{figs/fig_incomplete_observation.pdf}
%		\caption{The {incomplete and noisy observations} issue. The agent perceive information through sensors, while part of observation data is missing and is contaminated with noise.}\label{fig_incomplete_observation}
%	\end{figure}
%	\end{wrapfigure}
	In this paper we are interested in RL method under \emph{incomplete and noisy observation}. \emph{Incomplete} means part of observation is \emph{dynamically missing} over timestep, namely the {dimension} and {time} that observation missing occurs could not be known in advance by agent. 
%	While \emph{noisy} means that the data is contaminated with noise.

	A passive countermeasure against the incomplete observations is to stop the executing process. However, this arrangement suffers from high cost or even is dangerous under some circumstances, e.g., for an automatic drive car driving in high speed, stopping action when data missing occurs is fatal. \iffalse{For example, a Mars Exploration Robot, who is performing a task outside the airship by itself, should try its best to go back to the airship \ttt{only where they could be repaired}}.\fi
	Another typical solution is to fill the missing components with the adjacent earlier data. However, this method is trustless especially in a rapidly changing system. 
%	An alternative approach is to train models for each combination of components. By this way, the number of requisite models grows exponentially with the number of dimensions. 
	A plausible method is to predict values of missing components. However, since the environments involve noise, predicting the latent state accurately is a non-trivial task.
			
	Our work is part-in inspired by the approaches of POMDPs \citep{kaelbling1998planning, mcallister2017data}. During the execution phase of the agent, we maintain a \emph{belief state}, which is the posterior distribution over latent state based on the historical incomplete and noisy observations. The belief state is employed by the policy to make decision.
%	During the execution phase, we maintain a \emph{belief state}, which is a posterior distribution over latent state based on the incomplete and noisy observations. The belief state is updated through time and is employed by the policy to make decision. 
%	The updating process of belief state requires the transition model of MDP, which is estimated from the observations using a newly proposed surrogate loss function with local approximation. 
%	We test our method on several benchmark continuous control tasks, where we set the observation components is dynamically missing with a probability, while the other components are contaminated with additive noise.
%	We show that our algorithm can perform robustly against significant incompleteness and noise of observation, and learn better policy with fewer data than the original PPO.
%	\new{
	Our contributions are three-folds. 
	 First, we model the incomplete and noisy observations problem within the framework of POMDPs, and make several approximations to conduct a tractable and efficient belief updating.
	 Second, we propose a new surrogate loss function with local approximation, which learns the transition model of MDP from incomplete and noisy observations.
	 Last but not least, we propose a robust RL method with a belief imputation mechanism, which enables robust execution when the input is corrupted or partially missing. 
	 Extensive experiments on several benchmark tasks show that our approach outperforms several compared methods under various challenging scenarios.
%	}
%	\footnote{Videos are available at \url{https://sites.google.com/site/bipposupp/}}
%	Our main contribution is an RL method with belief imputation handling incomplete and noisy observations issues in RL, which we call \emph{Belief Imputation} PPO (\ProposedMethod[]), could learn from incomplete and noisy data. \ProposedMethod[] could perform robustly under a significant incompleteness and noise of observations.  Besides, the auxiliary task in belief imputation could also improve the original PPO by a large margin on performance and data efficiency and is competitive with the state-of-the-art model-free RL methods.
	
%	The paper is organized as follows. Section \ref{sec_Peliminaries} gives a brief outline of RL and related work. 
%%	Section \ref{sec_Peliminaries} introduces RL and PPO. 
%	Section \ref{sec_methods} details the new algorithm, \ProposedMethod[], and the generative model of the incomplete and noisy observations is presented in Section \ref{subsec_incomplete_observation}. Experiment results are demonstrated in Section \ref{sec_experiments}, and Section \ref{sec_conclusion} concludes paper.
	The remainder of the paper is organized as follows. First, we give a brief outline of related work, RL and the generative model. 
	Then we detail the new algorithm, \ProposedMethod[], and the generative model of the incomplete and noisy observations is presented. 
	After that, experimental results on RL benchmark tasks with incomplete and noisy observations are demonstrated. 
	The paper concludes with conclusions and directions for future work.

\section{Related Work}\label{sec_related_work}
	
	Concerning the methods handling missing data in RL. \cite{jette1997reinforcement} first present a data complement method by setting up an \emph{exception network} whose output is used to impute the missing values. Their method work in a MDP framework, and they impute missing data with the prediction based only on the imputed-state at last timestep. This differs from our approach, which works in a POMDP framework and employ the belief state to act based on the whole history observations and actions besides the current one.
	
	\cite{Lizotte2008} proposed a method for batch Q-learning with missing data. They model the posterior distribution of missing data $X_{miss}$ given observed data $X_{obs}$,
	where $X$ is a set of trajectories. 
	Then they sample $X^{(1)}_{miss},X^{(2)}_{miss},\ldots,X^{(N)}_{miss}$ from $X_{miss}|X_{obs}$ to construct multiple imputed datasets. The finally Q function is integrated with Q functions trained with these imputed data. Their method depends on proper assumption of the generative model and does not work well on high dimensional continuous control.
	%        \begin{align}
%			$\hat{Q}(s,a;X_{obs}) = \frac 1 N \sum_{i=1}^{N} Q(s,a;X_{obs},X^{(i)}_{miss}).$
	%        \end{align}
%			However, this method does not model the dynamics information of environment and works ineffective in high dimensional tasks due to sampling procedure.

%\new{ 
	For the methods working in POMDPs, the approach of maintaining a belief state is widely used. \cite{mcallister2017data} studied a special case where the partial observability has the form of additive Gaussian noise on the unobserved state. The belief is used to filter the noisy observations. \cite{igl2018deep} proposed the Deep Variational RL, which directly uses a network to output distribution of the belief state from the observation, and relies on a particle filter to approximate the intractable computation of the belief updating. \new{ Their method is more like a DNN-based 'black-box' which trying to directly output the belief state from current observation. While our method sufficiently exploits the prior knowledge of the problem structure under the incomplete and noisy observation setting by explicitly imputing the missing components from the observed part.}
	 
	\new{
	Many other approaches solve POMDPs by Recurrent Neural Networks (RNNs). 
	\cite{hausknecht2015deep} proposed the Deep Recurrent Q-network (DRQN), which employ RNNs to integrate historical trajectory and is robust to partial observability on Atari games.
	\cite{zhu2017improving} extended DRQN by explicitly including actions as input to the RNNs, which is called Action-specific DRQN (ADRQN). These works utilize RNNs to recurrently aggregate the history observations, while we maintain a belief state which is propagate forward by exploiting the problem structure.}
	
%	}
\section{Preliminaries}\label{sec_Peliminaries}
%	This work builds on earlier work from the RL community, specifically on the PPO actor-critic algorithm introduced in \cite{schulman2017proximal}. The rest of this section explains this prior work in more detail.
	\subsection{Reinforcement Learning}
%	\textbf{Reinforcement Learning.}
	In RL, the decision process is modelled as a Markov Decision Process (MDP) described by the tuple $M=(\mathcal{S},\mathcal{A},{\cal T},r,\rho_1,\gamma)$. $\mathcal{S}$, $\mathcal{A}$, ${\cal T}: \mathcal{S} \times \mathcal{A} \rightarrow \mathcal{P}(\mathcal{S})$, $r:\mathcal{S} \times{A} \rightarrow \mathbb{R} $ are the set of states, set of actions, stochastic transition function and the reward function respectively. $\rho_1$ is the distribution of the initial state $s_0$, and $\gamma \in (0,1)$ is the discount factor. 
	Value function $V^\pi(s)=\mathbb{E}[R_t^\gamma|s_t=s ]$ and action-value function $Q^\pi(s,a)=\mathbb{E}[R_t^\gamma|s_t=s,a_t=a ]$, where $R_t^\gamma=\sum_{k=0}^\infty \gamma^k r_{t+k+1}$ , are often used in RL algorithms.

%	In continuous control tasks, 
	The main idea of RL methods is to update the parameter of policy $\pi_\theta$ towards the direction of maximizing performance objective,
	\[L^{\textrm{policy}}(\theta ) = \int_{\cal S \times \cal A} {{\rho ^{{\pi _\theta }}}(s)} {\pi_{{\theta }}(a|s)r(s,a)dsda},\] where ${\rho ^{{\pi _\theta }}}(s) = (1-\gamma)\mathop \sum_{t = 1}^{\infty} {\gamma ^{t - 1}}{\rho_t^{{\pi _\theta }}}(s)$, $\rho_t^{\pi_\theta}(s)$ is the density of $s$ at time $t$. 
	The most commonly used gradient estimator has the form 
	\[
		\nabla_{\theta}L^{\textrm{policy}}(\theta) = \mathbb{E}_{s,a} \left[ {  \nabla_\theta \log \pi_\theta(a|s) A^{\pi_\theta}(s,a)} \right],
	\]
	where $A^{\pi_\theta}(s_t,a_t)=Q^{\pi_\theta}(s_t,a_t)-V^{\pi_\theta}(s_t)$ is the advantage function of policy $\pi_\theta$. The up to date algorithm to estimate $A^{\pi_\theta}(s_t,a_t)$ is the Generalized Advantage Estimator $\hat{A}^{(\gamma,\lambda)}_t$ \citep{Schulman2016HighDimensional}, which has the form
	\[
%	\begin{align}
	 \hat{A}^{(\gamma,\lambda)}_t=\sum_{k=0}^{\infty}{(\gamma\lambda)}^k\left[ r_t + \gamma \hat{V}_{\phi}(s_{t+k+1})-\hat{V}_{\phi}(s_{t+k}) \right],
%	 \label{eq_A_t}
%	\end{align}
	\]
	where $0<\lambda<1$ is a trade-off coefficient, and $\hat{V}_{\phi}$ is the estimated value function trained by 
	
	\begin{equation}
		{{L}^{\textrm{value}}({\phi})} = \mathbb{E}_{s_t} {\left\| {{V}_{\phi }}({{s}_t})-{{{\hat{V}}}_t} \right\|}^{2},
%		\label{eq_L_value}
	\end{equation}
	where $\hat{V}_t=\sum_{k=0}^{\infty}\gamma^k r_{t+k}$ is the discounted accumulated reward of from timestep $t$. 
%	The objectives is integrated to into an overall objective,
%	$
%		J(\theta ,\phi )= {{L}^\textrm{policy}({\theta})}+\lambda_{v}{{L}^\textrm{value}({\phi})}\label{eq_objective_ac}
%	$
%	, where  $\lambda_{v}$ is the coefficient.
%	\subsection{Proximal Policy Optimization}		
		PPO \citep{schulman2017proximal} tactfully plug the idea of constraint on policy into the objective, 
%		which originates from TRPO \cite{schulman2015trust}, by ignoring the change in probability ratio when it would make the objective improve too much. PPO optimize the "surrogate" objective
		by optimizing the "surrogate" objective
%		The right version
%		\begin{align}
%			${L}^\textrm{policy}_{\textrm{}}({\theta}) = \mathbb{E}_{s_t,a_t} \left[ \min \left( r_\theta(a_t,s_t){\hat{A}^{(\gamma,\lambda)}_t},\mathop{clip}\left( r_\theta(a_t|s_t),1-\delta ,1+\delta  \right){\hat{A}^{(\gamma,\lambda)}_{t}} \right) \right], \label{eq_L_policy}$
%		\end{align}
		\begin{dmath}
		{L}^{\textrm{policy}}_{\textrm{}}({\theta}) = \mathbb{E}_{s_t,a_t}
		\left[ \min \left( {\omega}_\theta(a_t,s_t){\hat{A}^{(\gamma,\lambda)}_t}, \\
		\mathop{clip}\left( {\omega}_\theta(a_t|s_t),1-\delta ,1+\delta  \right){\hat{A}^{(\gamma,\lambda)}_{t}} \right) \right],
		\end{dmath}
		where ${\omega}_\theta(a_t,s_t)=\frac{{{\pi }_{\theta }}({{a}_{t}}|{{s}_{t}})}{{{\pi }_{{{\theta }_{old}}}}({{a}_{t}}|{{s}_{t}})}$, $\delta$ is the clipping parameter. 
%		We choose PPO as a basis of our method for the following reasons. First, PPO is compatible with architectures that include parameter sharing, which means that it is possible to use our auxiliary task to assist in learning better representation. Second, it doesn't need experience replay, which decreases the cost of learning. Third, it is competitive with the state-of-the-art RL algorithms on several tasks.		
%		Fortunately, in RL, it is feasible to avoid observation data missing problem in training phase, since that we could simply repair or substitute the faulty sensors once we detect component missing. While put the agent into use, as discussed in section \ref{sec_intro}, the sensors could not be repaired or substituted immediately and have to respond to the environment in time with the incomplete missing data.
	\subsection{Generative Model of Incomplete and Noisy Observations}\label{subsec_incomplete_observation}
		Let the state at timestep $t$ to be $s_t \in \mathbb{R}^D$, which is noisily and partially observed as $x_t=  (s_t+\epsilon_t)\odot w_t$, where $\epsilon_t \sim \mathcal{N}(0,\Sigma^\epsilon)$, $w_t \in \{0,1\}^D$ is the observable indicator vector ($1$ for observed, $0$ for missing), and the operator $\odot$ is defined as
%		\begin{gather}
		$
			(s \odot w)^{(i)}\triangleq
			\left\{
			\begin{matrix}
			s^{(i)} & w^{(i)} =  1 \\
			\miss & w^{(i)} = 0
			\end{matrix}
			\right.,
		$
%		\end{gather}
		where a component possesses value $\miss$ means the data is missing; thus $x_t \in (\mathbb{R}\cup \{\miss\})^{D}$. 
		
		Note that the entries of $w_t$ are sampled at each timestep, which make the observation \emph{dynamically missing} over time. We refer \emph{missing part} the ones whose values are missing, denoted as $x_t^m \in \mathbb{R}^{M_t} $, $M_t$ is the number of components of the \emph{missing part}, and \emph{\loaded part} are the ones that possess values, denoted as $x_t^\ob \in \mathbb{R}^{D-M_t}$.
		Current RL systems could not deal with these incomplete observations directly, especially the case where the \emph{\loaded part} is dynamically changing over time. 
		
%		\new{
		%		i.e., $P(w_t|s_t,x_t \etparam)=P(w_t | \etparam)$. 
		%		For example, $w_t^{(i)} \sim \mathop{Bern}(1-\missparam^{(i)})$, $i=1,\ldots,D$, $\missparam^{(i)}$ is the probability of missing of component $i$ (and is same across timesteps), which we call the \emph{missing ratio}.
		The missing mechanism could be characterized by the conditional distribution $P(w_t|s_t,x_t \etparam)$, where $\missparam_t$ denotes unknown parameter that partly determine missingness. Our method works as long as $w_t$ is independent of $s_t$. Specifically, we assume that the observation data are either \emph{Missingness Completely At Random} (MCAR) or \emph{Missingness At Random} (MAR) \citep{little1987statistical}. 
	      MCAR means the missingness only depends on unknown parameters $\missparam_t$,
	      while MAR is less restrictive in that the missingness could also depend on other \loaded[] components, i.e.,$P(w_t|s_t,x_t \etparam)=P(w_t|x_t^\ob \etparam)$. Since MCAR case is a particular case of MAR, we would derive our method under the MAR case throughout the paper.
%		}
			
\section{Methods}\label{sec_methods}
	In this section, 
%	we first introduce the generative model of incomplete and noisy observations.
%	 in Section \ref{subsec_incomplete_observation}. 
	we present the modifications on two subroutines of RL methods, which are execution and training phase respectively.
	 During the execution phase, we conduct belief state propagation, which is employed by the policy to decide action.
%	 , detailed in Section \ref{subsec_execution}. 
	 While in training phase, we learn the transition model of MDP, which is required by updating of the belief state and jointly trained with policy searching.
%	 , detailed in Section \ref{subsec_training}.
%	The updating of belief state requires the transition function of MDP, thus in the training phase, we apply a DNN which we call \emph{prediction network} to learn the transition function from the incomplete observed data, detailed in Section \ref{subsec_training}. 
%	The execution phase and training phase are presented in Algorithm \ref{alg_main}.
	
	\subsection{Belief State Propagation}\label{subsec_execution}
		In the execution phase, {we divert the impracticable planning in original state space into \emph{belief space}.}
		In belief space, a \emph{belief state} is a posterior distribution over possible original state space based on the history trajectory data, i.e., $ b_t(s_t) \triangleq P(s_{t}|x_{1:t}, a_{1:t-1})$.
		The updating of belief state $b_t(s_t)$ has the form by using Bayes rule,
		\begin{equation}
			 b_{t}(s_{t})
			%			 &= \frac{ P(x_{t}|s_{t}) \left[ \int { b_{t-1}(s_{t-1}) {\cal T}(s_{t}|s_{t-1}, a_{t-1}) ds_{t-1}} \right] }{ \int{ P(x_{t}|s_{t}) \left[ \int { b_t(s_{t-1}) {\cal T}(s_{t}|s_{t-1}, a_{t-1}) ds_{t-1}} \right]} ds_{t} }
			 = \frac{ P(x_{t}|s_{t} \etparam) \tilde b_{t}(s_{t})}
			 {\int{ P(x_{t}|s'_{t} \etparam) \tilde b_{t}(s'_{t}) ds'_{t}}}, \quad \text{for } t=1,2,\ldots
			 \label{eq_transition}
		\end{equation}
		\begin{equation}
		\begin{aligned}
			\text{where } 
 			& \tilde b_{t}(s_{t})=\int { b_{t-1}(s_{t-1}) {\cal T}(s_{t}|s_{t-1}, a_{t-1}) ds_{t-1}} \\
 			& \text{for } t=2,3,\ldots \label{eq_transition_inter}
		\end{aligned}
		\end{equation}
		where ${\cal T}$ is the transition probability function in original state space. To clarify the updating process, we introduce $\tilde b_{t}(s_{t})$ which is called an "intermediate" belief state. 
		The execution starts with a prior intermediate belief state $\tilde b_1(s_1)$ and an initial observation $x_1$, the belief state $b_t(s_t)$ is computed using \eqref{eq_transition}, then the action $a_t$ is decided according to this belief state and receive an incomplete observation $x_{t+1}$, finally we yield the next belief state $ b_{t+1}$ by  computing \eqref{eq_transition_inter} and \eqref{eq_transition} alternatively, and keeps iterating until end of the episode. We detail this process in the following.
		
%		\textbf{Computing Intermediate Belief State.}
%		\new{
		First, we compute the intermediate belief state $\tilde b_{t}(s_{t})$ using \eqref{eq_transition_inter} (the initial intermediate belief state $\tilde b_1$ is a parameterized distribution), which is intractable since the transition probability function ${\cal T}(s_t|s_{t-1},a_{t-1})$ is nonlinear. However,
		\eqref{eq_transition_inter} could be approximated by replacing the nonlinear ${\cal T}(s_t|s_{t-1},a_{t-1})$ with first-order (linear) approximation of ${\cal T}$ w.r.t. $s_{t-1}$ at ${\mathbb E}_{s_{t-1} \sim b_{t-1}}[s_{t-1}]$, which reduces to $\tilde b_t(s_t) \doteq  {\cal T}(s_{t}| {\mathbb E}_{s_{t-1} \sim b_{t-1}}[ s_{t-1} ], a_{t-1} )$.
%		The first one is that, we approximate the nonlinear transition function ${\cal T}(s_t|s_{t-1},a_{t-1})$ using a first-order Taylor expansion about ${\mathbb E}_{s_{t-1} \sim b_{t-1}}[ s_{t-1} ]$, ${\mathbb E}_{s_{t-1} \sim b_{t-1}} \left[ {\cal T}(s_{t}| s_{t-1} , a_{t-1} ) \right] \doteq  {\cal T}(s_{t}| {\mathbb E}_{s_{t-1} \sim b_{t-1}}[ s_{t-1} ], a_{t-1} ) $.
%		$\bar s_{t-1}={\mathbb E}_{s_{t-1} \sim b_{t-1}}[ s_{t-1} ]$
%		In principle, the conditional transition probability ${\cal T}$ could be any distribution, e.g., Gaussian or Laplace, parametrized by functions of $s_{t-1}$ and $a_{t-1}$.
		For tractability, we approximate the transition function by using Laplace Approximation, formulated as $ {\cal T}(s_{t}|s_{t-1},a_{t-1})=\mathcal{N}\left( s_{t} | f^\mu(s_{t-1},a_{t-1}), f^\Sigma(s_{t-1},a_{t-1}) \right)$, where $f^\mu$ and $f^\Sigma$ is approximated by DNNs, detailed in next section. Thus, for an intermediate state space $\tilde b_t(s_t)$, the mean $\tilde \mu_t$ and variance $\tilde \Sigma_t$ are computed by
%		}
		
		\begin{dmath}
			\tilde b_{t}(s_{t}) 
%			&\doteq {\cal T}(s_{t}|{\mathbb E}_{s_{t-1} \sim b_{t-1}}[ s_{t-1} ], a_{t-1} ),a_{t-1})  \\
			\doteq \mathcal{N} \left( s_{t} | 
				{\tilde \mu_t = f^\mu\left( {\mathbb E}_{s_{t-1} \sim b_{t-1}}[ s_{t-1} ] ,a_{t-1}\right)}
				,\\
				{\quad \quad \quad \quad \tilde \Sigma_t = f^\Sigma\left( {\mathbb E}_{s_{t-1} \sim b_{t-1}}[ s_{t-1} ],a_{t-1}\right)}
			\right)
%			\tilde \mu_t = f^\mu\left( {\mathbb E}_{s_{t-1} \sim b_{t-1}}[ s_{t-1} ] ,a_{t-1}\right),
%			\tilde \Sigma_t = f^\Sigma\left( {\mathbb E}_{s_{t-1} \sim b_{t-1}}[ s_{t-1} ],a_{t-1}\right)
			\label{eq_transition_prior}
		\end{dmath}
%		\new{
		  Here we approximate the transition distribution as a multivariate Gaussian but focus on modeling the dynamic with a highly non-linear DNN, which helps to capture the complex dynamics of the environment. Although more general density models such as mixture of Gaussians or non-parametric models could be adopted, this could increase the difficulty of belief inference.
%		}
		
%		\textbf{Updating Belief State.}
		Then, the belief state $b_{t}$ is computed using \eqref{eq_transition}. For the likelihood part $P(x_{t}|s_{t} \etparam)$, where $x_{t} \in (R\cup \{\miss\})^D$ is an incomplete and noisy observation in which some components possess value $\miss$. \iffalse Note that the \loaded part is dynamically changing through time, therefore we first permute the variables to extract the \loaded components. \fi We denote $I^\ob_t=\{ i | x_t^{(i)} \neq \miss \}$, $I^m_t=\{ i | x_t^{(i)} = \miss \}$ as the \loaded and missing indexes respectively. A \emph{sub-permutation matrix}, which is used for filter and leave the \loaded part of $x_t$, is constructed by $\W_t \in \mathbb{R}^{(D-M_t) \times D}$ by $ \W_t^{(j, {I_t^{o}}^{(j)} )} = 1,j=1,\ldots,D-M_t$ while other entries are $0$. For example, for a $D=3$ observation $x_t=(1,\miss,2)^\top$, the sub-permutation matrix is 
		$
			\W_t = 
			\begin{pmatrix}
				1 & 0 & 0 \\
				0 & 0 & 1
			\end{pmatrix}
		$.
		\ttt{Following the generative model,} we have 
		\begin{equation}
		P(x_{t}|s_{t} \etparam)= P(w_t|x_t^a \etparam) P(x_{t}^\ob|s_{t})
		\label{eq_likelihood}
		\end{equation}
		where \new{$P(w_t|x_t^a \etparam)$ emerges due to the MAR assumption}, $P(x_{t}^\ob|s_{t})={\cal N}(\W_{t}x_{t}|\W_{t}s_t, \W_t \Sigma^\epsilon \W_t^\top)$ is marginal likelihood of the \loaded part $x_t^\ob$ and has the form of linear Gaussian model (derived in Appendix \ref{app-sec_ob_likelihood}). 
		
		In addition, we adopt Gaussian as the initial intermediate belief state $\tilde b_1(s_1)\triangleq{\cal N}(s_1|\tilde \mu_1, \tilde \Sigma_1)$. Thus by employing \eqref{eq_transition}, $b_1(s_1)$ is still Gaussian, and also does $\tilde b_2({s_2}), b_2(s_2), \ldots, \tilde b_t({s_t}), b_t(s_t)$ by employing \eqref{eq_transition_prior} and \eqref{eq_transition} alternatively. Specifically, given an intermediate belief state $\tilde b_t(s_t)={\cal N}(s_t|\tilde \mu_t, \tilde \Sigma_t)$, updating of belief state $b_t$ in \eqref{eq_transition} is computed by
%		$P(x_t|s_t)=\mathcal{N}\left( x_t| s_t, \Sigma^\epsilon \right)$

		\begin{dmath}
			{b}_{t}({s}_{t}) = \mathcal{N}\left( { s}_{t} | 
				{{\mu}_{t}\hiderel{=}{\tilde \mu _{t}} + {F_{t}}\left( {x_{t} - {\tilde \mu _{t}}} \right),}
			 \\{\quad \quad \quad \quad {\Sigma}_{t} \hiderel{=}  { \tilde \Sigma _{t}} - {F_{t}}{\tilde \Sigma _{t}} }
			  \right),
%				{\mu}_{t} = {\tilde \mu _{t}} + {F_{t}}\left( {x_{t} - {\tilde \mu _{t}}} \right),
%				{\Sigma}_{t} = { \tilde \Sigma _{t}} - {F_{t}}{\tilde \Sigma _{t}}
			\label{eq_posterior_belief}
		\end{dmath}
		\begin{flalign}
			&\text{where}& {F_{t}} = {\tilde \Sigma _{t}}\W_{t}^ \top {\left[ {{\W_{t}}({\tilde \Sigma _{t}} + \Sigma^\epsilon )\W_{t}^ \top } \right]^{ - 1}}{\W_{t}}. & \quad & \label{eq_F_t}
		\end{flalign}
		Note that the missing distribution $P(w_t|x_t^a \etparam)$ in \eqref{eq_likelihood} is cancelled out in \eqref{eq_transition} since it doesn't depend on state $s_t$. The derivation is detailed in Appendix \ref{app-sec_updating_belief}. 

%		\new{
		In some special cases, for example, when there is no data missing occurs (but noise exists), we have $\W_t=I$ and Eq \eqref{eq_F_t} reduces to  $F_t={\tilde \Sigma _{t}} {\left[ {({\tilde \Sigma _{t}} + \Sigma^\epsilon ) } \right]^{ - 1}}$. We could see that the larger the noise covariance $\Sigma^\epsilon$ the smaller the term $F_t$, which reduces the impact of current observation $x_t$ on the expectation $\mu_t$, and enlarges the uncertainty $\Sigma_t$ (see Eq \eqref{eq_posterior_belief}). 
		Furthermore, when the noise is also removed (and no data missing occurs), we have $\Sigma^\epsilon=\mathbf{0}$ and thus $F_t=I$, hence $\mu_t$ will collapse to $x_t$ and $\Sigma_t$ to $\mathbf{0}$.
%		}
%		To further understand the equation $\mu_t={\tilde \mu _{t}} + {F_{t}}\left( {x_{t} - {\tilde \mu _{t}}} \right)$, the term $\left( {x_{t} - {\tilde \mu _{t}}} \right)$ could be seen as a "rectificatory" term to the intermediate mean $\tilde \mu_t$, while $F_t$ describes the quantity to rectify. 
%		Seen in this light, in some special cases, for example, when there is no data missing occurs but noise exists, we have $\W_t=I$. Eq \eqref{eq_F_t} reflects that the larger the noise covariance $\Sigma^\epsilon$, the smaller the term $F_t$, which reduce the impact of rectification. 
%		Furthermore, when the noise also doesn't exist (and no data missing occurs), we extra have $\Sigma^\epsilon=\mathbf{0}$ and thus $F_t=I$, the mean $\mu_t$ will collapse to $x_t$, and the covariance $\Sigma_t$ to $\mathbf{0}$.
		
%		\textbf{Acting based on Belief State. }
		Finally, the belief state $b_t$ is employed by the policy $\pi$ to decide action, where we can use the exception $\mu_t$ of belief state for decision directly, 
%		\begin{equation}
		$
			\pi({b}_{t}) \triangleq  \pi( \mu_{t} ), \label{eq_act_posterior_mean}
		$
%		\end{equation}
		or include the uncertainty term $\Sigma_t$ as input,
		\begin{equation}
			\pi({b}_{t}) \triangleq  \pi( \mu_{t},\Sigma_t ), \label{eq_act_posterior_mean_1}
		\end{equation}
%		which we found to result in no significant advantage than just inputting the exception.
%		The overall procedure is presented in .
		
	\subsection{Learning Transition Model from Incomplete and Noisy Observations}\label{subsec_training}
%		In this section, we introduce methods to train the transition model.
%		initial intermediate belief state $\tilde b_1$, policy network and  \emph{prediction network} which is used to approximate transition function.
%		 which is required by updating of belief state. 
%		The policy and value network is trained by PPO with filtered imputed observations.
		The belief state propagation in \eqref{eq_transition_prior} requires a transition model, 
		and the initial intermediate belief state $\tilde b_1$ also need to be learned. 		
		\new{Some methods assume the transition model is already known \citep{Maxim2015deep} or is approximated by a generative network, which requires a lot of data to train especially in high-dimensional tasks.} However, we adopt Gaussian, which is parameterized by a highly non-linear \emph{transition network},  i.e., $ \hat{\mathcal  T}(s_{t+1}|s_t,a_t) = \mathcal{N}\left( s_{t+1} | f^\mu_{\psi_p}(s_t,a_t), f^\Sigma_{\psi_p}(s_t,a_t) \right)$. ${\psi_p}$ is parameter of the DNN, $f^\Sigma_{\psi_p}(s_t,a_t)$ is a state-action-dependent, positive-definite square matrix, which is parametrized by $f^\Sigma_{\psi_p}(s_t,a_t)=G_{\psi_p}(s_t,a_t)G_{\psi_p}^\top(s_t,a_t) $, where $G_{\psi_p}(s_t,a_t)$ is a lower-triangular matrix whose entries come from a linear output layer of a DNN. 
		The initial intermediate belief state is also a Gaussian distribution $\tilde b_1(s_1) \triangleq {\cal N}(s_1|\tilde \mu_1,\tilde \Sigma_1)$.
%		With old parameter $\psi'$ applied in $b_t(s_t; \psi')={\cal N}(s_t|\mu'_t,\Sigma'_t)$, the log-likelihood of observations $x_{1:T}$ given actions $a_{1:T}$ is 
%		The prediction network is employed to yield the next belief state, $b_{t+1}(s_{t+1})=\mathcal{N}\left( s_{t+1} | f^\mu_\psi(\hat s_t,a_t), f^\Sigma_\psi(\hat s_t,a_t) \right)$, where $\hat s_t$ is a filtered imputed state of last timestep from the posterior belief $\tilde b_t$. The next noisy incomplete observation data $x_{t+1}$ is a supervisory information to this prediction model. Thus the objective of prediction network could be modelled as log-likelihood loss of next observation $x_{t+1}$, which is a marginal distribution over the \loaded part $\W_{t+1}x_{t+1}$,
%		\begin{equation}
		For simplification, we denote $\psi \triangleq (\psi_p, \tilde \mu_1, \tilde \Sigma_1)$.
		The log-likelihood of observations given actions, i.e., $\log P(x_{1:T}|a_{1:T};\psi)$, is 
		\begin{equation}
		\begin{multlined}
			{L}(\psi) = 
				 \log  \int{ \tilde b(s_1;\psi)P(x_1|s_1 \etparam[1]) \cdot
				 } \\ {
				\prod_{t=1}^{T-1} \hat{\mathcal T}(s_{t+1}|s_t,a_t;\psi) P(x_{t+1}|s_{t+1} \etparam[t+1]) ds_{1:T}} \label{log_likelihood1}
		\end{multlined}
		\end{equation}
%		The log-likelihood $L(\psi)=\log \int{ P(s_1;\psi)P(x_1|s_1) \prod_{t=1}^{T-1} P(s_{t+1}|s_t,a_t;\psi) P(x_t|s_{t+1}) ds_{1:T}}$ is difficult to optimize directly, thus is substituted with the following local approximation, 
		{Note that the unknown parameter $\missparam_t$ (which partly determines the missing vector) doesn't need to be learned (is not employed by belief updating) and doesn't influence the learning of other parameters since it is independent of $s_t$ (see Eq \eqref{eq_likelihood}).}
		The multiple production in \eqref{log_likelihood1} makes the objective difficult to optimize directly. Instead, we introduce the following local approximation,
		\begin{dmath}
			{\hat L_{\psi'}}(\psi) = \log \int{ \tilde b_1(s_1;\psi)  P(x_1|s_1 \etparam[1])ds_1 } 
			\\
			\hiderel{+} \sum_{t=1}^{T-1} \left[ 
			\log  \int{ b_t(s_t;\psi') \hat{\cal T}(s_{t+1}|s_t,a_t; {\psi}) \cdot }
			\\
			{\quad \quad \quad  P(x_{t+1}|s_{t+1} \etparam[t+1])  ds_{t} ds_{t+1} } \right],
			\label{eq_approximate_likelihood}
		\end{dmath}
		where $\psi'$ is a value of variable $\psi$. 
%		TODO: Theorem: delete tmp
%		${\hat L_{\psi'}}$ matches $L$ to first order, formally we have
%		\begin{thm}
%			For any parameter $\psi'$,
%			\begin{align}
%				{\hat L_{\psi'}}(\psi')= L(\psi'), \quad \\
%				\left.\nabla_{\psi}{\hat L_{\psi'}}(\psi)\right|_{\psi=\psi'}= \left.\nabla_\psi L(\psi) \right|_{\psi=\psi'}
%				\label{eq_first_order_match}
%			\end{align}
%		\end{thm}
%		\noindent We provide proofs in Appendix \ref{app-sec_log_likelihood}. 
		
		Due to the intractable nonlinear $\hat {\cal T}$, Eq \eqref{eq_approximate_likelihood} is approximated by replacing the nonlinear $g(s_t, x_{t+1}, a_t)=\int{ \hat{\cal T}(s_{t+1}|s_t,a_t;{\psi}) P(x_{t+1}|s_{t+1}) ds_{t+1} }$ with first-order (linear) approximation w.r.t. $s_{t}$ at ${\mathbb E}_{s_t \sim b_{t}(s_t;{\psi'})}[s_t]=\mu_t$. Thus the final objective function has the form
%		\begin{equation}
		\begin{dmath}
				{\hat L}^{\textrm{model}}_{\psi'}(\psi) 
%					=&  \log \int{ \tilde b_1(s_1;\psi)  P(x_1|s_1)ds_1 } + \sum_{t=1}^{T-1} \left[ \log \int{ {\cal T}(s_{t+1}|\hat s_t,a_t; {\psi}) P(x_{t+1}|s_{t+1})  ds_{t} ds_{t+1} } \right] \notag \\
					= \log {\cal N} \left( \W_{1} x_{1} | \W_{1} \tilde \mu_1, \W_{1} \left( \tilde \Sigma_1 + \Sigma^\epsilon \right) \W_{1}^\top \right) 
					\\
					\hiderel{+} \sum_{t=1}^{T-1} \left[ \log 
								\mathcal{N}\left( \W_{t+1} x_{t+1} | \W_{t+1} f^\mu_{\psi_p}( \mu_t,a_t), \\
								{ \W_{t+1} \left( f^\Sigma_{\psi_p}( \mu_t,a_t) + \Sigma^\epsilon \right) \W_{t+1}^\top } \right)  \right]
		\end{dmath}
%		\end{equation}
%		where $\hat s_t={\mathbb E}_{s_t \sim b_{t}(s_t;{\psi'})}[s_t]$ is the mean of the belief state. 
		The parameters of the noise model could be learned in principle, e.g., by imposing extra regularization terms (e.g., low-rank constraints, $l_2$-norm) over the parameters. 
%		We experimented with the $l_2$-norm and achieved no significantly different results compared to the one with known noise.
%		By checking the part which employ $f^\Sigma_\psi(\hat s_t,a_t)$ and $\Sigma^\epsilon$ in \eqref{eq_posterior_belief} and \eqref{eq_L_predict_likelihood} , we can found that we don't have to separate this two terms which denote the variance of transition probability and the noise respectively, and just take the sum as a whole in execution and training. 
		
		\subsection{Jointly Learning with Policy Searching}
%		As for training policy, 
		We extend PPO from MDPs to a special-case of POMDPs (where the observations are incomplete and noisy), which we call \emph{Belief Imputation} PPO (\ProposedMethod[]). \new{The input to the policy and value network of PPO is modified to be belief state, i.e., $(\mu_t,\Sigma_t)$, as Eq \eqref{eq_act_posterior_mean_1} shows.} Whereas we only implement our method by extending PPO, it could also be generally extended to other DNN-based RL algorithms. 
%		We substitute the latent $s_t$ with $\hat s_t$, which is mean of the belief state, 
		We integrate the objectives by minimizing the following objective function,
		\begin{align}
			J(\theta ,\phi ,\psi )= -{{L}^{\textrm{policy}}_{\textrm{}}({\theta})}+\lambda_{v}{{L}^{\textrm{value}}({\phi})}-\lambda_{p}{{\hat L}^{\textrm{model}}_{\psi'}({\psi})} \label{eq_overall_objective}
		\end{align}
		where $\lambda_{v}$ and $\lambda_{p}$ are the coefficients. 
%		\new{
		The transition network shares parameter with the policy and value networks, and these three networks are trained jointly. 
		This architecture could assist in learning a more robust representation and promote the performance of RL, as we will evaluate in experiments. 
		Since we adopt the normalized advantage values and rewards to train policy and value network, all the three terms have the same magnitude across different tasks (the term ${{\hat L}^{\textrm{model}}_{\psi'}({\psi})}$ could be seen as a likelihood term). Thus we could use the same setting of the hyper-parameters across different tasks in principle.The \ProposedMethod[] algorithm is presented in Algorithm \ref{alg_main}.
%		}

		\begin{algorithm}[!h]
			\caption{\ProposedMethod[]}
			\begin{algorithmic}[1]\label{alg_main}
				\FOR{$i=1$ to TIMESTEPS\_MAX$/T$}
				\STATE \emph{// Execution Phase}
				\STATE Receive observation $x_1$ from environment
				\STATE Start with $ \tilde b_1=\mathcal{N}( \tilde \mu_1, \tilde \Sigma_1 )$
				\FOR{$t=1$ to $T$}
					\STATE Update the belief state $ b_t=\mathcal{N}( \mu_t,  \Sigma_t )$ by \eqref{eq_posterior_belief} %with $\tilde b_t$ and $x_t$
					\STATE Perform $a_t$ according to $\pi( b_t)$ by \eqref{eq_act_posterior_mean_1}
					\STATE Receive reward $r_t$ and incomplete observation $x_{t+1}$
					\STATE Update the intermediate belief state $\tilde b_{t+1}$ by \eqref{eq_transition_prior}
				\ENDFOR
				\STATE \emph{// Training Phase}
				\FOR {$k=1$ to Epoches\_Max}
					\STATE Compute advantage $\hat{A}_t^{(\lambda,\gamma)}$, for $t=1,\ldots,T$
					\FOR{$t=1$ to $T$}
						\STATE Update belief state $ b_t=\mathcal{N}( \mu_t,  \Sigma_t )$ by \eqref{eq_posterior_belief} with $\tilde b_t$ and $x_t$
%						, set $\hat s_t = \mu_t$
%						\STATE Set the filtered imputed state $\hat s_t= \mu_t$
						\STATE Train networks with the tuple $(\mu_t, \Sigma_t, a_t, A_t^{(\lambda,\gamma)}, x_{t+1})$ by \eqref{eq_overall_objective} %involving \eqref{eq_L_policy} \eqref{eq_L_value} \eqref{eq_L_predict_likelihood}
						\STATE Update the intermediate belief state $\tilde b_{t+1}$ by \eqref{eq_transition_prior}
					\ENDFOR   
				\ENDFOR
				\ENDFOR
			\end{algorithmic}
		\end{algorithm}

\section{Experiments}\label{sec_experiments}
	\subsection{Experiments Setting}		
		We designed our experiments to investigate the following questions:
		\begin{enumerate}
%			\new{
			\item Could \ProposedMethod be robust to the incomplete and noisy observations issue for control tasks? To what extent of missing and noise could it be robust? 
%			}
			\item Could \ProposedMethod contribute to learning regarding episode rewards and data efficiency? 
		\end{enumerate}
		
		To answer 1, we evaluate \ProposedMethod under different missing and noise settings on benchmark continuous control tasks. Specially, each component of observation has a probability of rate to be missing and is contaminated with additive Gaussian noise. 
		Concerning 2, we compare \ProposedMethod[] with several prior policy optimization algorithms, regarding episode rewards and the required timestep to hit a threshold. 
		The following two sections discuss these two questions respectively.

		\noindent \textbf{Simulated Tasks.}
		We evaluated our methods on 8 benchmarks simulated locomotion tasks, which is implemented in OpenAI Gym v0.9.3.
		\citep{Brockman2016OpenAI} using the MuJoCo \citep{Todorov2012MuJoCo} physics engine. The 8 benchmark tasks are HalfCheetah, Hopper, Walker2d, Ant, Swimmer, Reacher, InvertedDoublePendulum, InvertedPendulum respectively. 
		
		\noindent \textbf{Implementation Details.}
		We implemented our algorithms based on the implementations of PPO \citep{baselines}. Similarly in \cite{schulman2017proximal}, we used discount factor $\gamma=0.99$, GAE parameter $\lambda=0.95$, PPO clipping parameter is setting to be $0.2$, and Adam is used to for learning the weights of deep networks with a base learning rate of $lr=3\times10^{-4}$, Adam epsilon $\epsilon_{adam}=1\times10^{-5}$. Empirically, we set the penalty coefficient $\lambda_v=1.0, \lambda_p=1.0$. All three DNNs has two hidden layers of 64 units each and shares the first layer. 
%		\rewrite{Our code sets the random seeds for the random generators of all the workers and for all copies of the OpenAI Gym environments held by the workers.  To all methods on equal footing, we sampled three random seeds uniformly from the interval $[0,1000)$ and fixed them. Evaluation on three random seeds is widely adopted in literatures \citep{rajeswaran2017towards, schulman2017proximal}.}
	\subsection{Evaluation with Incomplete and Noisy Observations}\label{subsec_control_missing}
		\begin{figure*}[!b]
			\centering
			\includegraphics[width=1.0\textwidth]{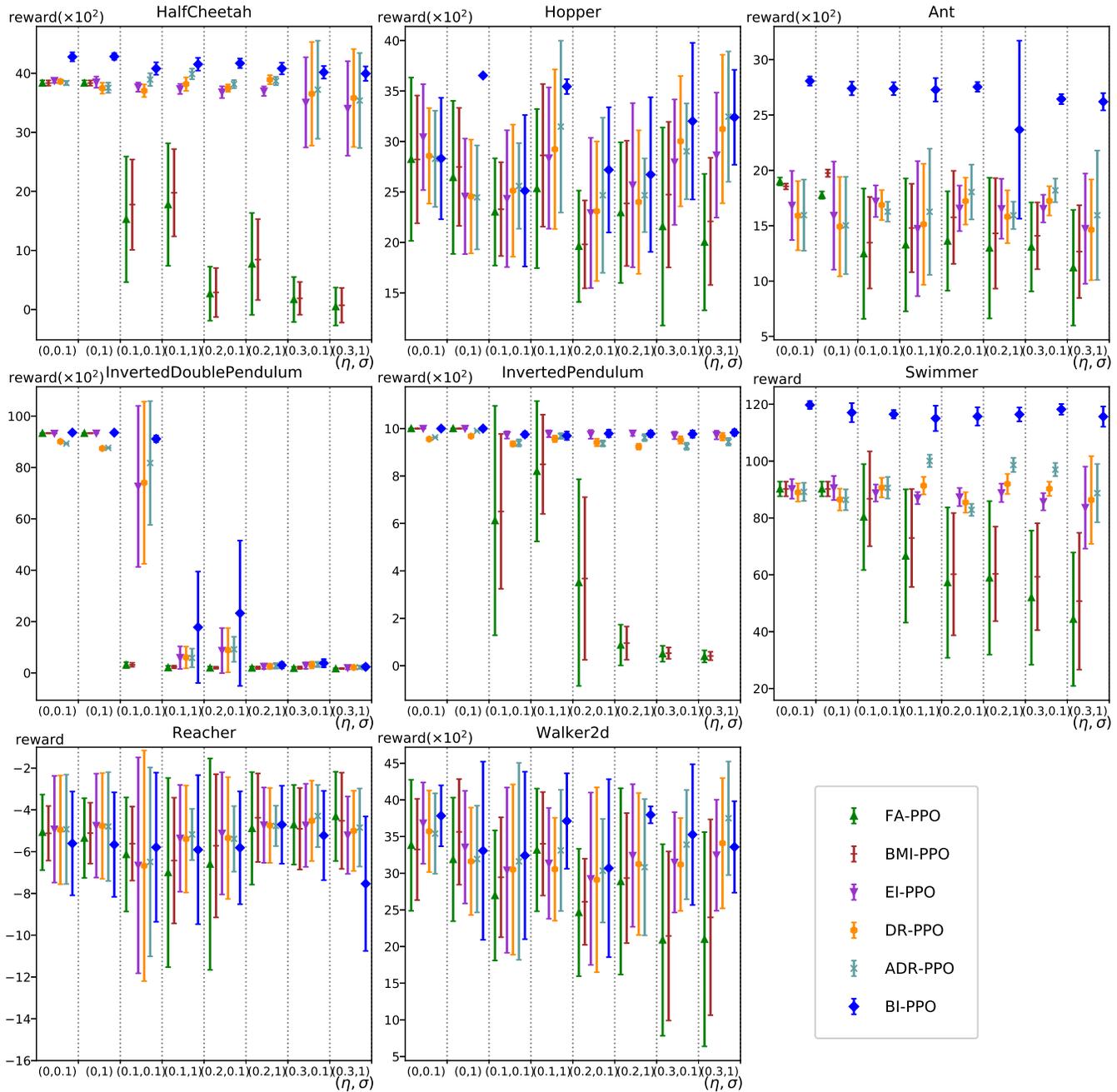}
			\caption{Episode rewards under different missing ratio $\eta$ and noise factor $\sigma$ settings on benchmark tasks. The error bars show $\pm 1$ standard deviation of the mean reward given 10 episodes of the trained model. The horizontal axis shows different combination of missing ratio $\eta$ and noise level $\sigma$, while the vertical axis shows the rewards. The rewards of different methods are shown for each setting. }\label{fig_rew}
		\end{figure*}
%		\begin{savenotes}
		\begin{table*}[!t]
			\centering
			\caption{Results of maximum attained episode rewards and timesteps to hit a threshold within 2 million timesteps, averaged over 3 random seeds. (a) shows the maximum attained episode reward. (b) shows the timesteps to hit a prescribed threshold. The thresholds for all environment were chosen according to \cite{wu2017scalable} except Ant and HalfCheetah since we evaluate algorithms within 1/15 of their limiting timesteps.}\label{tab_reward_hit}
%			\begin{tabular}{+p{3cm}<{\centering}^c|^c^c^c^c}
%			\begin{minipage}{\textwidth}
			\begin{tabular}{@{}
				+m{2.5cm}<{\centering}
				Ym{0.4in}<{\centering}Ym{0.4in}<{\centering}
				Ym{0.15in}<{\centering}Ym{0.36in}<{\centering}
				|Ym{0.38in}<{\centering}
				Ym{0.4in}<{\centering}Ym{0.4in}<{\centering}
				Ym{0.15in}<{\centering}Ym{0.36in}<{\centering}
				@{}}
				\toprule
%				& & \multicolumn{2}{c}{\ProposedMethod[]} & \multicolumn{2}{c}{EI-PPO}  &\multicolumn{2}{c}{PPO} & \multicolumn{2}{c}{ACKTR} \\% & \multicolumn{2}{c}{TRPO}
%				\rowstyle{} Task & Threshold & Rewards & Timesteps & Rewards & Timesteps & Rewards & Timesteps & Rewards & Timesteps \\
%				Task& & \multicolumn{4}{c}{Rewards} & \multicolumn{4}{c}{Timesteps} \\% & \multicolumn{2}{c}{TRPO}
				& \multicolumn{4}{c} {(a) Maximum Rewards}  & \multicolumn{5}{c}{(b) Timesteps ($\times 10^3$) to hit threshold }  \\
				\cline{ 2-5 }\cline{ 6\quad-10 }
				\noalign{\smallskip} 
				\rowstyle{\small} & \ProposedMethod[] & EI-PPO & PPO & ACKTR & Threshold & \ProposedMethod[] & EI-PPO & PPO & ACKTR \\
				\midrule{}
				%\multirow{8}{*}{$\epsilon=0.0$} 

				HalfCheetah                    & \textbf{4156}  &      3735      &      1549      &      3233      &      2100&  \textbf{235}  &      425       &$\infty$\footnotemark&      850       \\
				Hopper                         & \textbf{3828}  &      3595      &      3622      &      3309      &      2000&  \textbf{184}  &      231       &      209       &      502       \\
				Ant                            &      2988      &      1628      &      2448      & \textbf{3566}  &      2500&      1220      &    $\infty$    &    $\infty$    &  \textbf{825}  \\
				\small{Inverted DoublePendulum}& \textbf{9345}  &      9342      &      9333      &      9342      &      9100&  \textbf{104}  &      137       &      155       &      348       \\
				\small{InvertedPendulum}       & \textbf{1000}  &      1000      &      1000      &      1000      &       950&       18       &       22       &  \textbf{18}   &      110       \\
				Swimmer                        &      115       &      105       &  \textbf{118}  &       47       &        90&  \textbf{493}  &      1073      &      550       &    $\infty$    \\
				Reacher                        &       -5       &       -4       &  \textbf{-1}   &       -3       &        -7&      176       &      268       &  \textbf{137}  &      250       \\
				Walker2d                       & \textbf{4352}  &      4034      &      4285      &      2398      &      3000&      489       &      456       &  \textbf{252}  &    $\infty$    \\
				
				\bottomrule
				%\protect\footnotemark
				%\footnote{$\infty$ means that the method did not reach the reward threshold.}
			\end{tabular}
%			\end{minipage}
		\end{table*}
%		\end{savenotes}

		{
		To investigate robustness against incomplete and noisy observations, we investigate methods under different missing ratio and noise level. 
		For missing mechanism, we design for the MAR case, the probability function of $i$-th observable indicator ($i=1,\dots,D$) is
		\[
			P(w^{(i)}_t=1|x_t^a \etparam) \triangleq P( w^{(i)}_t=1|x_t^a, a_t, \missparam ) P(w^{(i)}_t=1|\missparam) 
		\]
%		where $P( w^{(i)}_t|x_t^a, a_t, \missparam )$ is a conditional distribution mainly depends on $x_t^a$ and $a_t$, $P(w^{(i)}_t|\missparam)$ is Bernoulli distribution. 
		The two terms in RHS represent the missing conditioned on observed and unknown variables, jointly impact $w_t^{(i)}$. Only both two distributions output $1$ could component $i$ be observable. They have forms
		\begin{equation*}
			\begin{gathered}
				\begin{multlined}
				P(w^{(i)}_t{=}1|x_t^a, a_t, \missparam)\triangleq 1- { min\left({g}({x_t^a}^\top\beta_{1,i} + a_t^\top\beta_{2,i}+ \beta_{3,i}), \missparam \right)}
				\end{multlined}
				\\
				P(w^{(i)}_t=1|\missparam) \triangleq 1-\missparam,
			\end{gathered}
		\end{equation*}
		where $g$ is the sigmoid function, $\beta_{1,i}$, $\beta_{2,i}$, $\beta_{3,i}$ are random variable sampled from Gaussian distribution, $0< \missparam < 1$ is the parameter of Bernoulli distribution.
		We use $\min$ operation to manually control the missing level in our experiments.
		 To explicitly compare the result under different missing level, we refer $\eta=\max_{x_t^a,a_t} P(w^{(i)}_t=0|x_t^a \etparam)$ to the \emph{missing ratio}, and we set $\eta=0,0.1,0.2,0.3$ for different settings by tuning $\missparam$.
		 The noise levels are controlled by a noise factor $\sigma=0.1,1$, such that the noises are $ \sqrt{ \Sigma^\epsilon }=\sigma \times 0.01 \times I$. We run each trial for $2\times 10^6$ timesteps (except for RNN-based comparison methods, which require more timesteps and is trained with $2\times10^7$ timesteps) and save the trained model every 2048 timesteps, then the model which achieves the best performance during the training process is employed to execute for 10 episodes. We measure the episode rewards, imputed state precision and execution time of each method. 
%		We first compare methods maintaining a deterministic predictive state or a belief state, then we compare the variants of our proposed method.
		}
		
%		\noindent \textbf{Comparative Methods.}
%		We compare five algorithms. To put the methods on equal footing, we implement these approaches by extending PPO. 
		We compare our method against the following algorithms. 1) The naive \emph{Fill Adjacent} (FA) method is baseline, where the missing value is imputed with adjacent earlier value. 
		2) Bayesian Multiple Imputation (BMI) method \citep{Lizotte2008}, which imputes missing values with posterior of observed data to build policies. 
		3) \emph{Exception Imputation} (EI) \citep{jette1997reinforcement}, which complement the missing values with a predictive state directly. 4) Deep Recurrent Q-network (DRQN) \citep{hausknecht2015deep}, which add recurrent LSTM architecture to the networks. %As the, the missing values are filled with $0$ and input to the RNNs.  
		5) Action-specific DRQN (ADRQN) \citep{zhu2017improving}, which extra including the action as input of RNNs. 
		We implement these methods using PPO, thus are called FA-PPO, BMI-PPO, EI-PPO, DR-PPO, ADR-PPO respectively.
		
%		Then we compare two variants of our proposed method, both of the which maintain a belief state during the execution phase, but update the belief and make decision based on different components of algorithm: 
%		4) \emph{Intermediate} \ProposedMethod[], which imputes the missing components with the mean of intermediate belief state (intermediate belief state). 5) \emph{Full} \ProposedMethod[], which employ the belief state during decision and transition process (full belief state).
		
		\noindent\textbf{Episode Rewards: }
		\fig\ref{fig_rew} shows episode rewards results under different missing ratio and noise level on benchmark tasks. At low noise without missing, all methods work relatively well. However, as missing ratio increase, FA-PPO works poorly, while BMI-PPO works slightly better than FA-PPO but a lot worse than \ProposedMethod[]. 
		DR-PPO, ADR-PPO performs better than EI-PPO, but much worse than BI-PPO. This is partly due to the reason that the hidden representation yielded by a RNN network is not stable when the input contains corrupted or missing values.
		\ProposedMethod[] outperform EI-PPO by a large margin, especially on HalfCheetah, Ant, Swimmer, Walker2d.
		All methods don't work well on InvertedDoublePendulum with high missing ratio. We speculate that the state in this task is compact and all components have critical information for decision.
		Nevertheless, overall \ProposedMethod[] shows robustness against significant incompleteness and noise, and work pretty well especially on high dimensional tasks like HalfCheetah, Hopper, and Ant.

		\noindent \textbf{Imputed State Precision: }
		The imputed state precision could partly reflect why the algorithms perform in the level.
%		since the more accurate imputation would provide more information for the agent to output more correct action. 
		We measure MSE between the imputed state and the latent system state, which could be obtained in the simulator. The imputed state precision results are almost consistent with rewards results in \fig\ref{fig_rew}, and are provided in Appendix \ref{app-sec_observation_precision}.
		
		\noindent \textbf{Execution Time Complexity: }
		We report model run time for EI-PPO, DR-PPO, ADR-PPO and \ProposedMethod[]. The former three ones run for average $2.7$s in an episode over all tasks, with constant time complexity w.r.t. the number of timetsteps $T$. While \ProposedMethod[] runs for average $5.1$s and scales with $\mathcal{O}\left(T(D(1-\eta))^3\right)$ time complexity. 
	
	\subsection{Evaluation with Complete and Clean Observations}\label{subsec_performance}

		In this section, we investigate whether \ProposedMethod could assist in learning better policy, regarding rewards and sample efficiency. Thus we evaluate them with complete and clean observations.
		We compared \ProposedMethod and EI-PPO against the following methods: the original PPO \citep{schulman2017proximal}, Actor Critic using
		Kronecker-Factored Trust Region (ACKTR) \citep{wu2017scalable}. Both of \ProposedMethod[] and EI-PPO set up an additional \emph{transition network}, but approximate the transition model in a different way.
%		but \ProposedMethod[] models the transition function of MDP as probabilistic while EI-PPO models as deterministic. 
		We use OpenAI Baselines \citep{baselines} implementations of these algorithms. Each algorithm runs for $2$ million timesteps, and is averaged over the 3 random seeds.
		\footnotetext{$\infty$ means that the method did not reach the reward threshold within $2$ million timesteps.}
		
		\noindent \textbf{Maximum Episode Rewards: }
		Table \ref{tab_reward_hit} (a) shows maximum episode rewards within 2 million timesteps. Table \ref{tab_reward_hit} (a) shows that both \ProposedMethod and EI-PPO significantly outperform the original PPO. Especially on HalfCheetah and Ant, \ProposedMethod achieves $268\%$ and $120\%$ of the original PPO maximum reward. This shows that our approach of jointly learning the required model for computing the belief state with the policy is beneficial to improve the overall generalization capability of the system.
		On HalfCheetah, Hopper, InvertedDoublePendulum, and Walker2d, \ProposedMethod performs better than ACKTR. \ProposedMethod outperform EI-PPO on almost all tasks except Reacher, which shows that the probabilistic approximation to transition function is necessary.
		
		\noindent \textbf{Sample Efficiency:}
		Table \ref{tab_reward_hit} (b) shows the timesteps required by algorithms to hit a prescribed threshold within $2$ million timesteps. The thresholds for all environments were chosen according to \cite{wu2017scalable} except Ant and HalfCheetah since we evaluate algorithms within 1/15 of their timesteps. Table \ref{tab_reward_hit} (b) shows that \ProposedMethod significantly outperform ACKTR on HalfCheetah, Hopper, InvertedDoublePendulum, Reacher. We could also observe that \ProposedMethod requires fewer timesteps than EI-PPO and the original PPO on almost all tasks except Swimmer and Reacher.

%	In this paper, we also employ this idea for imputation but deal with more complex issues, i.e., the observations are incomplete.
\section{Conclusions}\label{sec_conclusion}
%Our contribution is three-fold. First, we construct a generative model of incomplete and noisy observations and approximately update the belief state within the framework of POMDP. Second, we propose a newly surrogate loss function with local approximation, which could learn transition model of MDP from incomplete and noisy observations. Third, by combining the former two ones, we propose a robust RL method with belief imputation, which enables robust execution even under a significant incompleteness and noise. 

%We addressed the issue within the framework of partially observable Markov Decision Process (POMDP) using a model-based method, in which the transition model is estimated from the incomplete noisy observations using a newly proposed surrogate loss function with local approximation, while the policy and value function is learnt with the help of belief imputation. For the later purpose, a generative model is constructed and is seamlessly incorporated into the belief updating procedure of POMDP, which enables robust execution even under a significant incompleteness and noise. The effectiveness of the proposed method is verified on a collection of benchmark tasks, showing that our approach outperforms several compared methods under various challenging scenarios.
In this paper, we propose a new algorithm, \ProposedMethod[], addressing the incomplete and noisy observations problem for continuous control using a model-based method, in which the transition model is estimated from the incomplete and noisy observations using a newly proposed surrogate loss function with local approximation. 
To implement belief imputation for execution and training, we construct a generative model and seamlessly incorporate it into the belief updating procedure of POMDP. Furthermore, we propose a robust RL method with belief imputation, which enables robust execution even under a significant incompleteness and noise. 
Experiments verify the effectiveness of the proposed method, showing that our approach outperforms several compared methods under various challenging scenarios. Besides, \ProposedMethod[] could also improve the original PPO by a large margin regarding rewards and data efficiency, and is competitive with the state-of-the-art policy gradient methods.

%\new{
As part of our future work, we plan to handle a slightly different but more difficult case where some sensors suddenly produce totally different values. An approach to handle this case, based on this work, is to treat such values as outliers, and weight down their influence on the belief imputation or just throw them away.
%}
%	During the execution phase, we maintain a belief state which transits through timesteps and is employed by the policy to make decision. Furthermore, we propose methods to train PPO from incomplete and noisy observation data, and an additional prediction network is trained for maintaining more precise belief state.
%\subsubsection*{Acknowledgments}

\bibliographystyle{aaai}
\bibliography{IncompleteObservation}

\clearpage


\section*{Supplementary material}

## A Execution phase

### A.1 Observation Likelihood

We give the derivation of observation likelihood $P(x_t|s_t, a_t; \zeta_t)$ here. First we would repeat the probability model of observation. The state of the system $s_t$ is noisily and partially observed as $x_t = (s_t + \epsilon_t) \odot w_t$, where $\epsilon_t \sim \mathcal{N}(0, \Sigma^\epsilon)$, $w_t \in \{0,1\}^D$ is the observable indicator vector, $x_t \in (\mathbb{R} \cup \{\mathsf{o}\})^D$, in which o means the data is missing. We make the MAR assumption that $P(w_t|s_t, x_t, a_t; \zeta_t) = P(w_t|x_t^a, a_t; \zeta_t)$. Denote $x_t^m \in \mathbb{R}^{M_t}$ as the *missing part* whose values are missing, where $M_t$ is the number of the *missing part*, and $x_t^a \in \mathbb{R}^{D-M_t}$ as the *available part* that possess values. Besides, $I_t^a = \{i|x_t^{(i)} \neq \mathsf{o}\}$, $I_t^m = \{i|x_t^{(i)} = \mathsf{o}\}$ are the available and missing indexes respectively, $\mathsf{H}_t \in \mathbb{R}^{(D-M_t) \times D}$ is the *sub-permutation matrix*. The observation likelihood is

$$P(x_t|s_t, a_t; \zeta_t) = P(x_t^a, w_t|s_t, a_t; \zeta_t) \tag{1}$$
$$= P(w_t|s_t, x_t^a, a_t; \zeta_t) P(x_t^a|s_t)$$
$$= P(w_t|x_t^a, a_t; \zeta_t) P(x_t^a|s_t) \tag{2}$$

(1): We could construct missing vector $w_t$ by $w_t^{(i)} = \mathbb{1}\{x_t^{(i)} \neq \mathsf{o}\}$.

### A.2 Updating Belief State during Execution

Following the approximations in the paper for computing intermediate belief state $\tilde{b}_t(s_t) = \int b_{t-1}(s_{t-1}) \mathcal{T}(s_t|s_{t-1}, a_{t-1}) ds_{t-1}$, we have

$$\tilde{b}_t(s_t) \doteq \mathcal{N}\left(s_t|\tilde{\mu}_t = f^\mu\left(\mathbb{E}_{s_{t-1} \sim b_{t-1}}[s_{t-1}], a_{t-1}\right), \tilde{\Sigma}_t = f^\Sigma\left(\mathbb{E}_{s_{t-1} \sim b_{t-1}}[s_{t-1}], a_{t-1}\right)\right). \tag{3}$$

The belief state is updated by

$$b_t(s_t) = \frac{P(x_t|s_t, a_t; \zeta_t) \tilde{b}_t(s_t)}{\int P(x_t|s_t', a_t; \zeta_t) \tilde{b}_t(s_t') ds_t'} \tag{4}$$

Following the probability model of incomplete and noisy observation in the paper, we have $P(x_t|s_t) = P(w_t|x_t^a, \eta_t) P(x_t^a|s_t)$, where $P(x_t^a|s_t) = \mathcal{N}(\mathsf{H}_t x_t|\mathsf{H}_t s_t, \mathsf{H}_t \Sigma^\epsilon \mathsf{H}_t^\top)$ is the likelihood of the available part $x_t^a$, $\eta^{(i)}$ is the missing ratio of component $i$, $I_t^m$ and $I_t^a$ are indexes of missing part and available part respectively, $W_t \in \mathbb{R}^{(D-M_i) \times D}$ is the permutation matrix, $D$ and $M_i$ are dimension and number of available part respectively. Thus (4) could be written as

$$b_t(s_t) = \frac{P(w_t|x_t^a, a_t; \zeta_t) P(x_t^a|s_t) \tilde{b}_t(s_t)}{\int P(w_t|x_t^a, a_t; \zeta_t) P(x_t^a|s_t') \tilde{b}_t(s_t') ds_t'}$$
$$= \frac{P(x_t^a|s_t) \tilde{b}_t(s_t)}{\int P(x_t^a|s_t') \tilde{b}_t(s_t') ds_t'} \tag{5}$$

Since $P(x_t^a|s_t) = \mathcal{N}(\mathsf{H}_t x_t|\mathsf{H}_t s_t, \mathsf{H}_t \Sigma^\epsilon \mathsf{H}_t^\top)$, and given an intermediate belief state $\tilde{b}_t(s_t) = \mathcal{N}(s_t|\tilde{\mu}_t, \tilde{\Sigma}_t)$ which is also Gaussian distribution, thus updating of belief state $b_t$ in (5) could be derived from the conditional distribution of linear Gaussian model,

$$b_t(s_t) = \mathcal{N}\left(s_t|\mu_t = \tilde{\mu}_t + F_t(x_t - \tilde{\mu}_t), \Sigma_t = \tilde{\Sigma}_t - F_t \tilde{\Sigma}_t\right), \tag{6}$$

where $F_t = \tilde{\Sigma}_t \mathsf{H}_t^\top \left[\mathsf{H}_t(\tilde{\Sigma}_t + \Sigma^\epsilon)\mathsf{H}_t^\top\right]^{-1} \mathsf{H}_t$.

## B Experiment: Imputed State Precision

The imputed state precision could partly reflect why the algorithms perform in the level since the more accurate imputation would provide more information for the agent to output correct action. We measure the mean squared error of the imputed state between the latent system state, which could be obtained in the simulator. Fig. 1 shows the results. BI-PPO has a lower error than EI-PPO methods on most of the tasks, and the imputed state is robust against significant incompleteness and noise observations. The results in Fig. 1 are almost consistent with rewards results showed in the main content.

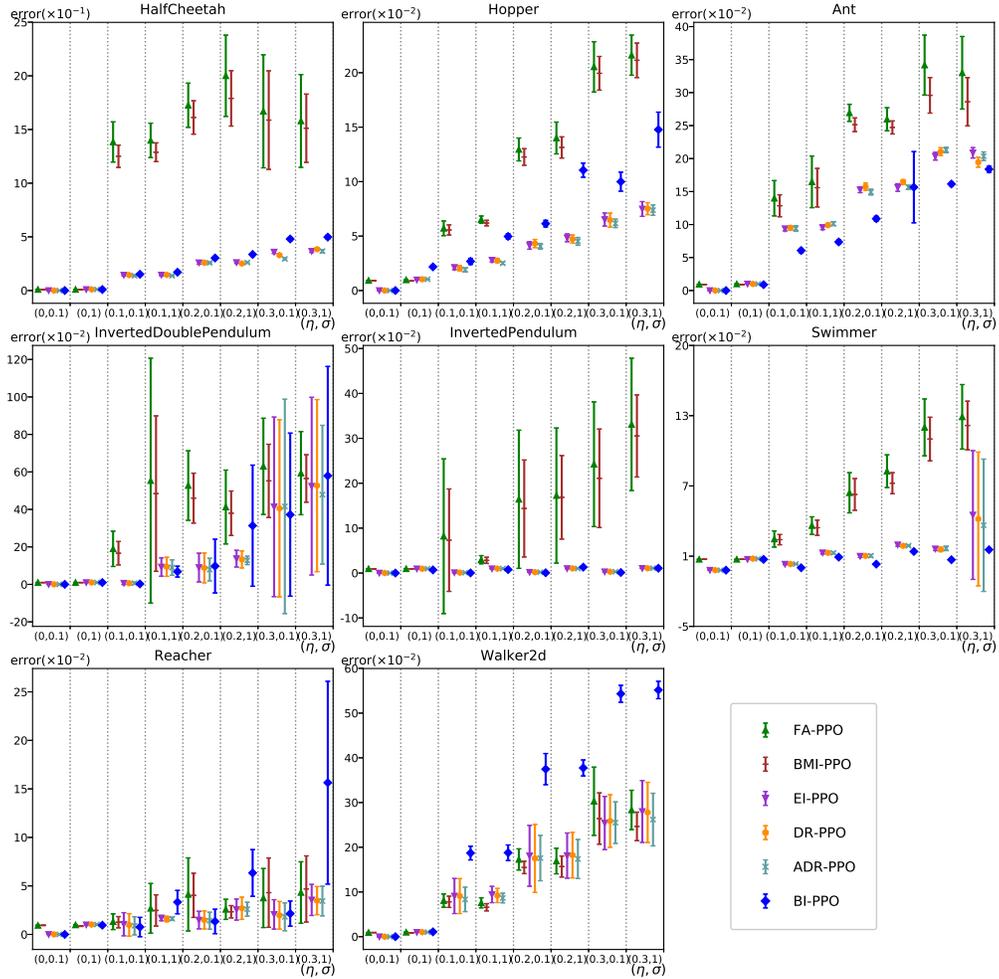

Figure 1: The MSE between imputed state and latent state under different missing ratio $\eta$ and noise factor $\sigma$ settings on benchmark tasks. The error bars show $\pm 1$ standard deviation of the mean squared error given 10 episodes of the trained model. The horizontal axis shows different combination of missing ratio $\eta$ and noise level $\sigma$, while the vertical axis shows squared error of (filtered) imputed observation between latent state. The squared error of different methods are shown for each setting.



\end{document}